\documentclass[letterpaper]{article} 
\usepackage{aaai23}  
\usepackage{times}  
\usepackage{helvet}  
\usepackage{courier}  
\usepackage[hyphens]{url}  
\usepackage{graphicx} 
\usepackage{natbib}  
\usepackage{caption} 
\usepackage{multirow}
\usepackage{booktabs}
\usepackage{algpseudocode}
\usepackage{color}
\usepackage{amsmath}
\usepackage{amsfonts}
\usepackage{wrapfig}
\usepackage{subcaption}
\usepackage{lipsum}
\usepackage[ruled,vlined,algo2e]{algorithm2e}
\usepackage{verbatim}

\usepackage{algorithm}

\usepackage{newfloat}
\usepackage{listings}
\DeclareCaptionStyle{ruled}{labelfont=normalfont,labelsep=colon,strut=off} 
\floatstyle{ruled}
\newfloat{listing}{tb}{lst}{}
\floatname{listing}{Listing}
\title{Dream to Generalize: Zero-Shot Model-Based Reinforcement Learning \\for Unseen Visual Distractions}

\author {
    Jeongsoo Ha\textsuperscript{\rm 1}, 
    Kyungsoo Kim\textsuperscript{\rm 2}, 
    Yusung Kim\textsuperscript{\rm 3}
}
\affiliations {
    
    \textsuperscript{\rm 1} Mechatronics Research, Samsung Electronics\\
    \textsuperscript{\rm 2} Intelligent Agent Lab, NCSOFT\\
    \textsuperscript{\rm 3} Department of Computer Science and Engineering, Sungkyunkwan University\\
    js1210.ha@samsung.com, unigary@ncsoft.com, yskim525@skku.edu
}

\begin{document}

\maketitle

\begin{abstract}
Model-based reinforcement learning (MBRL) has been used to efficiently solve vision-based control tasks in high-dimensional image observations. Although recent MBRL algorithms perform well in trained observations, they fail when faced with visual distractions in observations. These task-irrelevant distractions (e.g., clouds, shadows, and light) may be constantly present in real-world scenarios. In this study, we propose a novel self-supervised method, \textbf{Dr}eam to \textbf{G}eneralize (\textbf{Dr. G}), for zero-shot MBRL. \textbf{Dr. G} trains its encoder and world model with dual contrastive learning which efficiently captures task-relevant features among multi-view data augmentations. We also introduce a recurrent state inverse dynamics model that helps the world model to better understand the temporal structure. The proposed methods can enhance the robustness of the world model against visual distractions. To evaluate the generalization performance, we first train \textbf{Dr. G} on simple backgrounds and then test it on complex natural video backgrounds in the DeepMind Control suite, and the randomizing environments in Robosuite. \textbf{Dr. G} yields a performance improvement of 117\% and 14\% over prior works, respectively.
Our code is open-sourced and available at \url{https://github.com/JeongsooHa/DrG.git}

\end{abstract}

\section{Introduction}

Reinforcement learning (RL) with visual observations has achieved remarkable success in many areas, including video games, robot control, and autonomous driving~\cite{mnih2013playing, levine2016end, nair2018visual, kalashnikov2018qt, andrychowicz2020learning}. Because learning a control policy from high-dimensional image data is inevitably more difficult than learning from low-dimensional numerical data, training a visual RL agent requires a larger amount of training data. 
To address the data inefficiency, recent model-based RL (MBRL) studies have proposed learning a world model in the latent space, followed by planning the control policy in the latent world model. 

Although latent-level MBRL studies have successfully improved data efficiency, they have inherent drawbacks because they are typically designed as reconstruction-based methods.
The drawback of these methods comes from visually distracting elements (task-irrelevant information) that can compromise the accuracy of the reconstruction-based representation learning.
In particular, task-irrelevant information such as clouds, shadows, and light may change continuously depending on the time and place of the test. 
Therefore, generalization in terms of representation and policy learning is crucial for solving real-world problems~\cite{kim2022SPD}.  

\begin{figure}[t]
\centering
\includegraphics[width=0.45\textwidth]{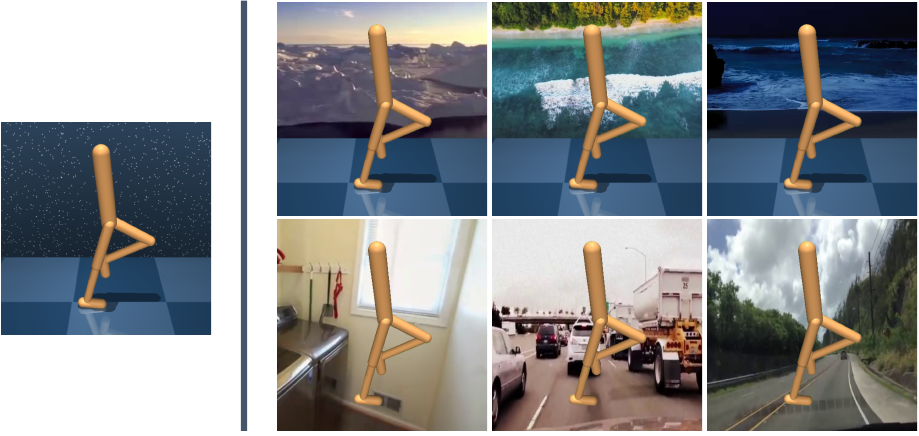}
\caption{The agent is trained in a simple background environment on the DeepMind Control suite in the default setting (left). We demonstrate the generalization ability in unseen environments in video easy settings (a top row on the right) and video hard settings (a bottom row on the right).}
\label{fig:train_test}
\end{figure}

In this study, we propose \textbf{Dr}eam to \textbf{G}eneralize (\textbf{Dr. G}), a \textit{zero-shot} MBRL framework that can be robust to visual distractions not experienced during training. Our proposed \textbf{Dr. G} introduces two self-supervised methods; 1) \textit{Dual Contrastive Learning (DCL)}, and 2) \textit{Recurrent State Inverse Dynamics (RSID)}.
\textbf{Dr. G} uses the same structure as the recurrent state space model (RSSM)~\cite{hafner2019learning} also used in Dreamer~\cite{hafner2019dream} but replaces the reconstruction-based learning part with DCL. 
The DCL approach consists of two objective functions over multi-view data augmentations. One objective function is applied between realities, which are latent states encoded with different data augmentation techniques (hard and soft) for the same image observation. It improves the generalization ability of the encoder against visual distractions. The other objective function is applied between reality and dreams (imagined latent states by RSSM). This allows the world model to dream (predict) the next latent state more robustly, enabling \textbf{Dr. G} to learn a more generalized control policy in the world model dreams.

\noindent The second self-supervised method, RSID, infers the actual executed actions over a sequence of latent states imagined by the world model. It enables the world model to understand the temporal structure and relationships between successive states, and helps to generate more robust rollouts for policy planning. 

\noindent We evaluate the generalization performance of the proposed zero-shot MBRL framework, \textbf{Dr. G}, on six continuous control visual tasks in the DeepMind Control suite~\cite{tassa2018deepmind} and on five tasks in the Robosuite~\cite{zhu2020robosuite}. After training \textbf{Dr. G} on simple background observations, we test it on unseen complex visual distractions, as shown in Figure~\ref{fig:train_test}. \textbf{Dr. G} yields a performance improvement of 117\% over existing model-based and model-free RL algorithms on the DeepMind Control suite and 14\% over existing algorithms on the Robosuit.

\begin{figure*}[t]
\begin{center}
\includegraphics[width=0.9\textwidth]{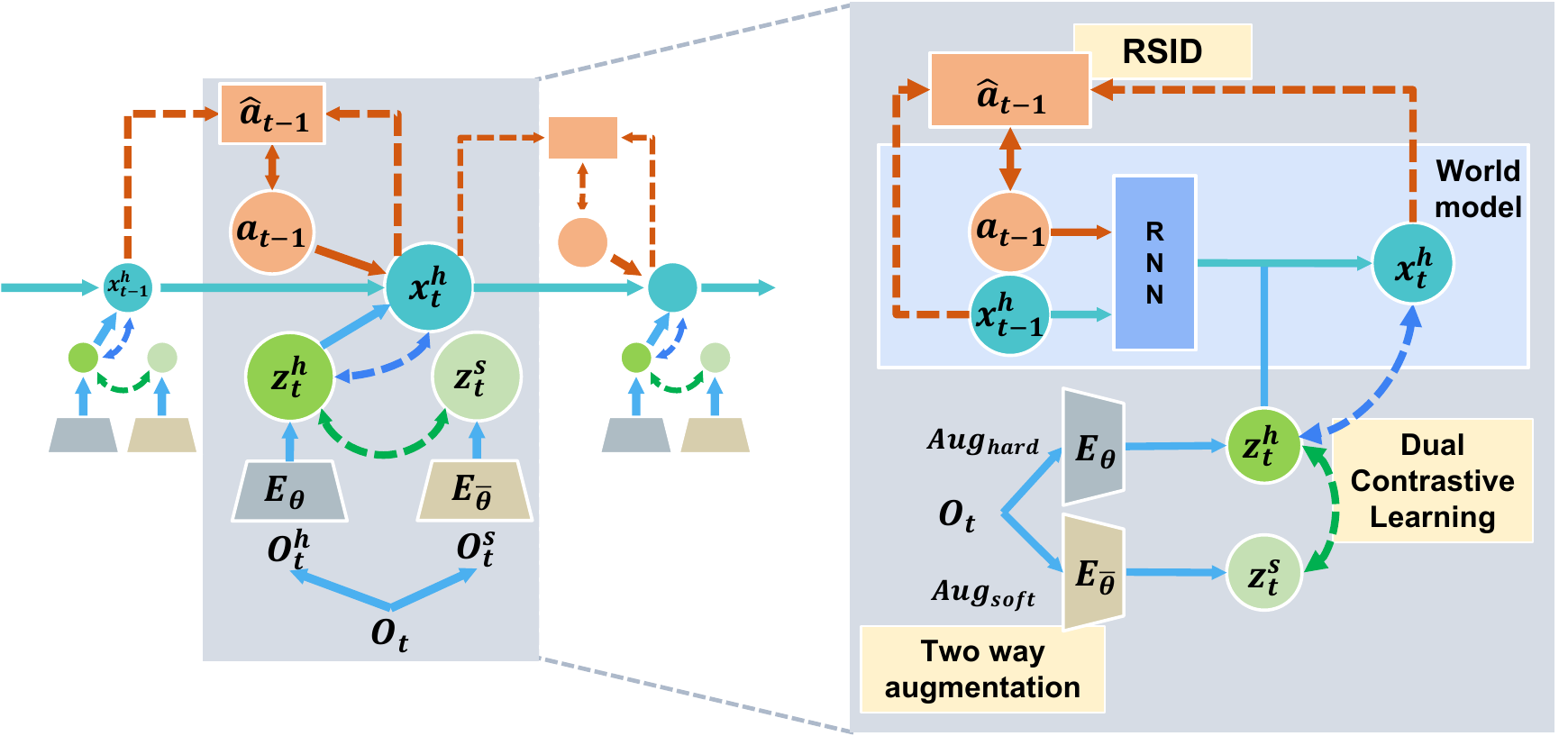} 
\end{center}
\caption{
Our Framework Overview: 
\textbf{Dr. G} trains the encoder and world model through \textit{two-way augmentation}, \textit{Dual Contrastive Learning} (green and blue dashed line), and \textit{Recurrent State Inverse Dynamics} (orange dash line) with sequential data. 
}

\label{fig:overview}
\vspace{0.3cm}
\end{figure*}

The key contributions of this study are as follows:
\begin{itemize} 

\item We introduce a zero-shot MBRL method, \textbf{Dr. G}, to train both the encoder and world model in a self-supervised manner. Using DCL and RSID, \textbf{Dr. G} can achieve robust representations and policies over unseen visual distractions. 
\item We demonstrate that \textbf{Dr. G} outperforms prior model-based and model-free algorithms on various visual control tasks in the DeepMind control suite and Robosuite. We also conduct thorough ablation studies to analyze our proposed method.
\end{itemize}

\section{Preliminaries}
In this section, we briefly introduce the training method for the world model, which forms the core of MBRL. 
The reconstruction-based world model is based on Dreamer~\cite{hafner2019dream} and Dreamerv2~\cite{hafner2020mastering}. 
For convenience, as the frameworks of the two papers are similar, we omit the version of Dreamerv2.

\subsection{Reconstruction-Based World Model Learning}

Recent MBRL methods train compact latent world models using high-dimensional visual inputs with variational autoencoders (VAE)~\cite{kingma2013auto} by optimizing the \textit{Evidence Lower BOund} (ELBO)~\cite{bishop2006pattern} of an observation. 
For an observable variable $x$, VAEs learn a latent variable $z$ that generates $x$ by optimizing an ELBO of $\log p(x)$, as follows:
\begin{equation} 
\label{eq1}
\begin{aligned}
&\makebox[-0.8cm]{\hfill}\log p(x) = \log\int p(x|z)p(z)dz\\
&\makebox[0.5cm]{\hfill}\ge \mathbb{E}_{q(z|x)}[\log p(x|z)]-D_{\mathrm{KL}}[q(z|x)||p(z)],
\end{aligned}
\end{equation}
where $D_{KL}[q(z|x)||p(z)]$ represents the Kullback-Leibler divergence between the prior distribution $p(z)$ and an assumed distribution $q(z|x)$ that samples $z$ conditioned on $x$. 

Dreamer~\cite{hafner2019dream}uses RSSM~\cite{hafner2019learning} as the world model to predict the sequence of future states and reward signals in latent space. 
At each time step $t$, the agent receives an image observation $o_t$ and a reward $r_t$ (from $a_{<t}$ in the sequential decision-making task). 
Then, the agent chooses an action $a_t$ based on its policy. 
RSSM learns latent dynamics by reconstructing images and rewards by optimizing the ELBO of $\log p(o_{1:T},r_{1:T}|a_{1:T})$~\cite{hafner2019learning, igl2018deep}. 
That is, as RSSM optimizes ELBO for sequential information, as expressed in Equation~\ref{eq1}, we obtain 
\begin{equation} 
\label{eq2}
\begin{aligned}
&\log p(o_{1:T}, r_{1:T}|a_{1:T})\\
& = \log\int\prod^T_{t=1}p(o_t|s_{\le t}, a_{< t})p(r_t|s_{\le t}, a_{< t})\\
&\makebox[4.5cm]{\hfill}p(s_t|s_{< t}, a_{< t})ds_{1:T} \\
& = \log\int\prod^T_{t=1}p(o_t|h_t, s_t)p(r_t|h_t, s_t)p(s_t|h_t)ds_{1:T},
\end{aligned}
\end{equation}
where $s_{1:T}$ are sequential states in the stochastic model, and $h_t$ is the hidden state vector obtained through $f(h_{t-1}, s_{t-1}, a_{t-1})$ as a deterministic state, which uses gated recurrent unit (GRU)~\cite{cho2014properties}.
To infer the agent states from past observations and actions, a variational encoder is used, which is expressed as

\begin{equation} 
\label{eq3}
\begin{aligned}
&\makebox[-1.9cm]{\hfill}q(s_{1:T}|o_{1:T},a_{1:T}) = \prod^T_{t=1}q(s_t|s_{< t}, a_{<t}, o_t) \\
&\makebox[0.7cm]{\hfill}= \prod^T_{t=1}q(s_t|h_t,o_t)
\end{aligned}
\end{equation}

Based on Equations~\ref{eq2} and~\ref{eq3}, the objective of Dreamer is to maximize the ELBO, as follows.
\begin{equation} 
\label{eq4}
\begin{aligned}
& \mathcal{J}_\mathrm{Dreamer}= \sum^T_{t=1}\mathbb{E}_q [ \log p(o_t|h_t,s_t)+\log p(r_t|h_t,s_t) \\
&\makebox[2.5cm]{\hfill} - D_{\mathrm{KL}}(q(s_t|h_t,o_t)||p(s_t | h_t))]\\
&\makebox[1.5cm]{\hfill}= \sum^T_{t=1}\mathbb{E}_q \left[ \mathcal{J}^t_\mathrm{O} + \mathcal{J}^t_\mathrm{R} - \mathcal{J}^t_{\mathrm{KL}} \right]
\end{aligned}
\end{equation}
$\mathcal{J}_O$ and $\mathcal{J}_R$ are used as reconstruction objective functions to restore image observations and rewards. 
And $\mathcal{J}_{KL}$ is used as an objective function for the KL divergence.

\subsection{Improvement Actor and Critic With Latent Dynamics}
After training RSSM as a world model, the actor and critic are trained through latent trajectories imagined in latent space using a fixed world model.
On imagined trajectories with a finite horizon $H$, the actor and critic learn behaviors that consider rewards beyond the horizon. 
At this time, the reward and the next state are predicted by the trained world model.
At each imagination step $\tau \ge t$, during a few steps $H$, the actor and critic are expressed as follows:

\begin{description} 
    \setlength\itemsep{0.5em}
    \item{\makebox[2cm]{Actor model:\hfill} $\hat{a}_\tau \sim \pi_\phi(\hat{a}_\tau | h_{\tau}, s_{\tau})$}
    \item{\makebox[2cm]{Critic model:\hfill} $v_\psi(s_\tau) \approx \mathbb{E}_{q(\cdot|s_\tau)}(\sum^{t+H}_{\tau=t} \gamma^{\tau-t}r_\tau)$},
\end{description}

where $\theta$, $\psi$ are model parameters and $\gamma$ is a discount factor.
The actor and critic are trained cooperatively in policy iterations. 
A policy model aims to maximize a value estimate, whereas a value model aims to match the value estimate to a behavioral model. 
Within the imagined trajectory, the actor and critic are trained to improve the policy such that $\lambda$-return~\cite{sutton2018reinforcement, schulman2015high} and approximate $\lambda$-return using squared loss are maximized, respectively.
\subsection{Contrastive Learning}
Contrastive learning~\cite{hadsell2006dimensionality, he2020momentum, wu2018unsupervised, chen2020simple, oord2018representation} is a framework for learning representations that satisfy similarity constraints in a dataset typically organized based on similar and dissimilar pairs. 
It can be understood as training an encoder for a dictionary look-up task. 
It considers an encoded query $q$ and a set of encoded samples $\mathbb{K}=\{ k_0, k_1, k_2, ... \}$ as the keys of a dictionary. 
Assuming that there is a single key (denoted as $k_+$) in the dictionary that matches $q$, the goal of contrastive learning is to ensure that $q$ matches $k_+$ to a greater extent than the other keys in $\mathbb{K} \backslash \{k_+\}$ (except a single sample $k_+$ in a set $\mathbb{K}$). 
$q, \mathbb{K}, k_+$ and $\mathbb{K} \backslash \{k_+\}$ are referred to as the anchor, target, positive, and negative, respectively, in the terms of contrastive learning~\cite{oord2018representation, he2020momentum}. 
Similarities between the anchor and targets are best modeled by calculating the dot product ($q^Tk$)~\cite{wu2018unsupervised, he2020momentum} or bilinear products ($q^TWk$)~\cite{oord2018representation, henaff2020data}. 
To learn embeddings that satisfy the similarity relations, CPC~\cite{oord2018representation} proposed the InfoNCE loss, which is expressed as

\begin{equation} 
\label{eq5}
\resizebox{.9\hsize}{!}{$\mathcal{L}_q = \log \frac{\exp (q^TWk_+)}{\exp{(q^TWk_+)}+ \sum^{K-1}_{i=0}\exp{(q^TWk_i)}}$}
\end{equation}
The loss Equation~\ref{eq5} can be understood as the log-loss of a $K$-way softmax classifier whose label is $k_+$~\cite{laskin2020curl}.

\section{Method}
We propose \textbf{Dr. G}, a novel self-supervised method, to train \textit{zero-shot} MBRL. 
Dr. G achieves excellent generalization ability for observational changes not experienced during training.
The proposed approach is illustrated in Figure~\ref{fig:overview}.

\subsection{Model Overview}

The basic architecture is based on the Dreamer~\cite{hafner2019dream} paradigm. 
We first train RSSM as the world model and then plan the control policy on the rollouts imagined by the world model. 
The actor and critic learning methods used are the same as in Dreamer; however, we replace the reconstruction objective of Dreamer with the proposed self-supervised methods, namely \textit{DCL} and \textit{RSID}. 
In the self-supervised methods, we apply multi-view data augmentations-a soft augmentation that provides minor position changes and a hard augmentation that inserts complex visual distractors to interfere with the original image. 
The combination of multi-view data augmentations and the self-supervised methods successfully achieve zero-shot generalization.
We show the overall training process through the Algorithm~\ref{algo}.

\subsection{Hard and Soft Augmentations}
\begin{figure}[t]
\centering
\includegraphics[width=0.42\textwidth]{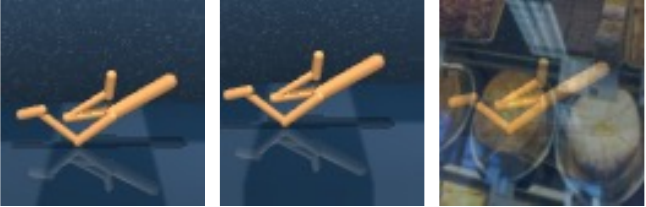}
\caption{Random shift and random overlay are used as soft and hard augmentations, respectively. Original image observation (left), soft augmented version (mid), and hard augmented version (right).}
\label{fig:augmentation}
\end{figure}

We use two data augmentation techniques during the encoder and world model training; one is \textit{Random-Shift}~\cite{kostrikov2020image} as a soft augmentation $Aug_s$ and the other is \textit{Random-Overlay}~\cite{hansen2020self, hansen2021generalization} as a hard augmentation $Aug_h$ as shown in Figure~\ref{fig:augmentation}. 
Random shift applies a pad around the observation image and performs a random crop back to the original image size. 
Random overlay linearly interpolates between the original observation $o_t$ and a randomly chosen complex image, as in Equation~\ref{eq6}, where $\alpha$ is the interpolation coefficient and $D$ is a complex image dataset containing 1.8M diverse scenes~\cite{hansen2021generalization}.

\begin{equation} 
\label{eq6}
\begin{aligned}
& o^s_t = \mathrm{Aug_s}(o_t) \overset{\Delta}{=} \mathrm{Random\ shift}(o_t)\\
& o^h_t = \mathrm{Aug_h}(o_t) \overset{\Delta}{=} (1-\alpha)*o_t+\alpha*\mathrm{img},\ \mathrm{img}\sim D
\end{aligned}
\end{equation}

\subsection{Dual Contrastive Learning}

Instead of the reconstruction loss used in Dreamer, we introduce \textit{DCL} using InfoNCE loss~\cite{oord2018representation}, which is a widely used loss function in contrastive model training. 
Our DCL consists of two objectives: One is applied between latent states with multi-view augmentations for the same observation. 
It enables the encoder to extract invariant representations (task-relevant features) regardless of different augmentations (dominated by task-irrelevant information). 
Because these latent states are encoded from real observations, we call it contrastive learning between realities. 
The other objective is applied between reality (latent states encoded from the real observations) and dreams (latent states generated by the world model), making the world model more robust.

\SetKwInput{Kwpara}{Hyperparameters}
\SetKwInput{Kwinit}{Initialize}
\newcommand\mycommfont[1]{\footnotesize\ttfamily{#1}}
\SetCommentSty{mycommfont}

\begin{algorithm}[t]
\DontPrintSemicolon
  
\Kwpara{\\
S(seed episode),
C(collect interval),
B(batch size),\\
L(sequence length),
H(imagination horizon)\\
}
\Kwinit{dataset $D$ with $S$ random seed episodes.}
\Kwinit{neural network parameters $\theta, \phi, \psi$.}
\For{each iteration}
{

    \For{updata setp $c=1..C$}
    {
        Draw $B$ data sequences $ {(o_t, a_t, r_t)}^{k+L}_{t=k} \sim D$\\
        \tcp*{Update world model $\theta$}
        Train the world model $\mathbb{E}_B[J_{Dr. G}(z, x, a|\theta)]$\\
        \tcp*{Update actor $\phi$ and critic $\psi$}
        
        Train the actor $\mathbb{E}_B[J_{\pi_\phi}]$\\
        Train the critic $\mathbb{E}_B[J_{Q_\psi}]$
    }
    
    \tcp{Interaction with real environment}
    $o_1 \sim $ \texttt{env.reset}(), $x_0 \leftarrow 0$, $a_0 \leftarrow 0$\\
    \For{environment step $t=1..T$}
    {
        $z_t \leftarrow E_\theta(o_t)$\\
        $x_t \leftarrow \mathrm{RSSM}_\theta(z_t, x_{t-1}, a_{t-1})$\\
        $a_t \sim \pi_\psi(a_t|x_t)$\\
        $r_t, o_{t+1} \leftarrow$ \texttt{env.step}($a_t$)\\
        $D \leftarrow D \cup(o_t, a_t, r_t)$
    }
 }   
\caption{Dr. G}
\label{algo}
\end{algorithm}

\begin{table*}[t] 
\centering
\renewcommand{\arraystretch}{1.2}
\resizebox{1\textwidth}{!}{
\renewcommand{\arraystretch}{1.1}
\begin{tabular}{c|c|c|c|c|c|c|c|c|c}
    \toprule
    \textbf{Setting} & \textbf{Task} & Dr. G (ours) & SAC & CURL & PAD & SODA & SECANT & Dreamer & DreamerPro \\
    \midrule
    \midrule
    \multirow{6}{*}{\textbf{\textit{Video easy}}}
    & Ball in cup Catch & 701$\pm$36 & 172$\pm$46 & 316$\pm$119 & 436$\pm$55 & 875$\pm$56 & \textbf{903$\pm$49} & 90$\pm$87  & 56$\pm$55\\
    & Cartpole Swingup  & 572$\pm$25 & 204$\pm$20 & 404$\pm$67 & 521$\pm$76 & \textbf{758$\pm$62} & 752$\pm$38 & 120$\pm$42  & 174$\pm$69\\
    & Cheetah Run       & \textbf{547$\pm$21} ({$+$27\%}) & 80$\pm$19 & {151$\pm$16} & 206$\pm$34 & 220$\pm$10 &  428$\pm$70 & 48$\pm$30  & 41$\pm$16 \\
    & Hopper Hop        & \textbf{191$\pm$28} ({$+$154\%}) & 56$\pm$21 & 10$\pm$15 & 67$\pm$5  & 75$\pm$29 & - & 38$\pm$11 & 28$\pm$8 \\
    & Walker Run        & \textbf{449$\pm$63} ({$+$71\%}) & 79$\pm$2 & {253$\pm$11} & 210$\pm$8  & 262$\pm$12 & -& 60$\pm$22 & 84$\pm$49 \\
    & Walker Walk       & \textbf{902$\pm$23} ({$+$7\%}) & 104$\pm$14 & {556$\pm$33} & 717$\pm$79 & 768$\pm$38 & 842$\pm$47 & 1$\pm$2  & 24$\pm$15 \\
    \midrule
    \multirow{6}{*}{\textbf{\textit{Video hard}}}
    & Ball in cup Catch & \textbf{635$\pm$26} ({$+$94\%}) & 98$\pm$25 & 115$\pm$33 & 66$\pm$61 & 327$\pm$100 & - & 74$\pm$73 & 103$\pm$100 \\
    & Cartpole Swingup & \textbf{545$\pm$23} ({$+$27\%}) & 165$\pm$8 & 114$\pm$15 & 123$\pm$24 & 429$\pm$64 & - & 126$\pm$29 & 158$\pm$49 \\
    & Cheetah Run     & \textbf{489$\pm$11} ({$+$237\%}) & 81$\pm$12 & 18$\pm$5     & 17$\pm$8 & 145$\pm$14 & - & 28$\pm$4 & 32$\pm$14 \\
    & Hopper Hop      & \textbf{181$\pm$19} ({$+$212\%}) & 11$\pm$9 & 2$\pm$2       & 4$\pm$3 & 58$\pm$15 & - & 25$\pm$5 & 37$\pm$9 \\
    & Walker Run      & \textbf{421$\pm$39} ({$+$242\%}) & 59$\pm$3 & 49$\pm$3      & 51$\pm$2 & 123$\pm$21 & - & 57$\pm$18 & 44$\pm$14 \\
    & Walker Walk     & \textbf{782$\pm$37} ({$+$105\%}) & 49$\pm$14 & 58$\pm$18    & 93$\pm$29 & 381$\pm$72 & - & 1$\pm$1 & 3$\pm$2 \\
    \bottomrule
\end{tabular}
}
\caption{Performance of \textbf{Dr. G} and baselines on six tasks in the DeepMind Control suite. We evaluated the trained model in video easy and video hard settings. \textbf{Dr. G} outperforms state-of-the-art baselines by an average of 117\% on 10 out of 12 tasks. Each task was run with three seeds.}
\label{result_table}
\end{table*}

\subsubsection{Contrastive Learning Between Realities} 
In \textit{reality-reality}, the agent compares latent states embedded in the encoder $E_\theta$ and target encoder $E_{\bar\theta}$ to improve the generalization ability against task-irrelevant information, as indicated by the green dashed line in Figure~\ref{fig:overview}. 
We consider the latent state encoded from the observation as \textit{reality}. 
The encoder embeds a hard augmented observation in $z^h$, and at the same time the target encoder extracts another latent state $z^s$ with a soft augmented version. 
To maximize the mutual information of two latent states with different perspectives, we use contrastive learning and an objective function of the following form:

\begin{equation} 
\label{eq7}
\begin{aligned}
\resizebox{.90\hsize}{!}{$\makebox[-0.2cm]{\hfill}\mathcal{J}^t_{\mathrm{E}} = \mathbb{E} \left[ \log \frac{\exp({z^s_t}^T W {z^h_{t,+}})}{\exp({z^s_t}^T W {z^h_{t,+}}) + \sum^{N-1}_{i=0}\exp({z^s_t}^T W {z^h_{t,i}})} \right]$}
\end{aligned}
\end{equation}
where $z^h_t= E_\theta(\mathrm{Aug}_h(o_t))$ and $ z^s_t = E_{\bar\theta}(\mathrm{Aug}_s(o_t))$, which apply hard and soft augmentation to image observation respectively. 
We apply soft and hard augmentation to each of the N images (observations) in the batch.
For each soft-augmented latent state, the hard-augmented latent state for the same image is used as a positive sample, and the other N-1 hard-augmented versions are used as negative samples.
Then the target encoder is updated according to the following \textit{Exponential Moving Average} (EMA): 

\begin{equation} 
\label{eq8}
\begin{aligned}
\bar{\theta}_{n+1} \leftarrow (1-\tau)\bar{\theta}_n+\tau\theta_n,
\end{aligned}
\end{equation}
for an iteration step $n$ and a momentum coefficient $\tau \in (0,1]$, such that only parameters $\theta$ are updated by gradient descent~\cite{he2020momentum, grill2020bootstrap, lillicrap2015continuous}.

\subsubsection{Contrastive Learning Between Dream and Reality}
In \textit{dream-reality}, shown as the blue dashed line in Figure~\ref{fig:overview}, the agent compares \textit{dream} $x^h_t$, which is a latent state imagined by the world mode, with \textit{reality} $z^h_t$ encoded from the augmented observation. 
By maximizing the similarity between dream and reality, the world model can imagine (predict) the next latent state more robustly. 
We note that \textit{dream-reality} uses the hard augmentation technique only because it shows the best zero-shot generalization empirically. 
This ablation study is shown in Figure 10 in the supplementary material. 
By updating in a similar way to \textit{reality-reality}, contrastive learning takes the following form: 

\begin{equation} 
\label{eq9}
\begin{aligned}
\resizebox{.90\hsize}{!}{$\makebox[-0.2cm]{\hfill}\mathcal{J}^t_{\resizebox{0.03\textwidth}{!}{RSSM}}\makebox[-0.1cm]{\hfill}= \mathbb{E}\left[\log \makebox[-0.1cm]{\hfill} \frac{\exp({z^h_t}^T W {x^h_{t,+}})}{\exp({z^h_t}^T W {x^h_{t,+}}) + \sum^{N-1}_{i=0}\exp({z^h_t}^T W {x^h_{t,i}})} \right]$}
\end{aligned}
\end{equation}
where $x^h_t = \mathrm{RSSM}_\theta(z^h_t, x^h_{t-1}, a_{t-1})$ which is hard augmented imagined latent state, $z^h_t = E_{\theta}(\mathrm{Aug}_h(o_t))$ is the encoded latent state from hard augmented image observation.
We use N reality and dream states in the batch. For each reality state, the dream state for the same image is used as a positive sample, and the other N-1 dream states are used as negative samples.

\subsection{Recurrent State Inverse Dynamics}

The goal of \textit{RSID} is to improve the robustness of the imagination by allowing world models to better understand the dynamics of tasks. 
The world model needs to generate a series of imagined latent states from the initial latent state, which is encoded from the observation. 
Because we input the hard augmented observation to improve robustness, understanding the relationship between successive states proves a challenge. 
To address this, we let the world model learn the causal relationship between successive imagined latent states by inferring the actual executed action. 
RSID can infer actions $\hat a_{t}$ from the imagined latent states $x^h_{t} = \mathrm{RSSM}_\theta(z^h_{t}, x^h_{t-1}, a_{t-1})$ obtained during training RSSM, as follows: 
\begin{equation} 
\label{eq10}
\begin{aligned}
\hat a_{t} = \mathrm{RSID}_{\theta}(x^h_{t}, x^h_{t+1}).
\end{aligned}
\end{equation}
From the imagined latent state $x^h_t$, the actions inferred via $\mathrm{RSID}_\theta$ are trained to be similar to the actual performed actions using MSE loss: 
\begin{equation} 
\label{eq11}
\begin{aligned}
\mathcal{J}^t_{\mathrm{RSID}} = \mathbb{E}\left[ \mathrm{mse}(a_t, \hat a_t) \right].
\end{aligned}
\end{equation}

Finally, we combine the previous objective functions to obtain the proposed objective function, which enables the world model and encoder training for generalization to yield appropriate policies for complex image observation. 
The proposed objective function is defined as 
\begin{equation} 
\label{eq12}
\begin{aligned}
\resizebox{.87\hsize}{!}{$\mathcal{J}_{\mathrm{Dr. G}} = \sum^T_{t=1}\mathbb{E}[\mathcal{J}^t_{\mathrm{E}}+\mathcal{J}^t_{\mathrm{RSSM}}+ \mathcal{J}^t_{\mathrm{RSID}}+\mathcal{J}^t_\mathrm{R}-\mathcal{J}^t_{\mathrm{KL}}]$}
\end{aligned}
\end{equation}
$\mathcal{J}_\mathrm{R}$ and $\mathcal{J}_{\mathrm{KL}}$ are terms to reconstruct the reward and compute KL divergence, respectively, and are same as those of Dreamer~\cite{hafner2019dream} objective functions. 

\section{Experiments}

We compare the zero-shot generalization performance of our method with the current best model-free and model-based methods on six continuous control tasks from the Deepmind Control suite (DMControl~\cite{tassa2018deepmind}) and Robosuite~\cite{zhu2020robosuite}. 
All methods were trained with default simple backgrounds but evaluated with complex backgrounds. 
Finally, we demonstrate the importance of each combination component in our method through ablation studies.

\begin{figure*}[t]
\centering
\includegraphics[width=1.0\textwidth]{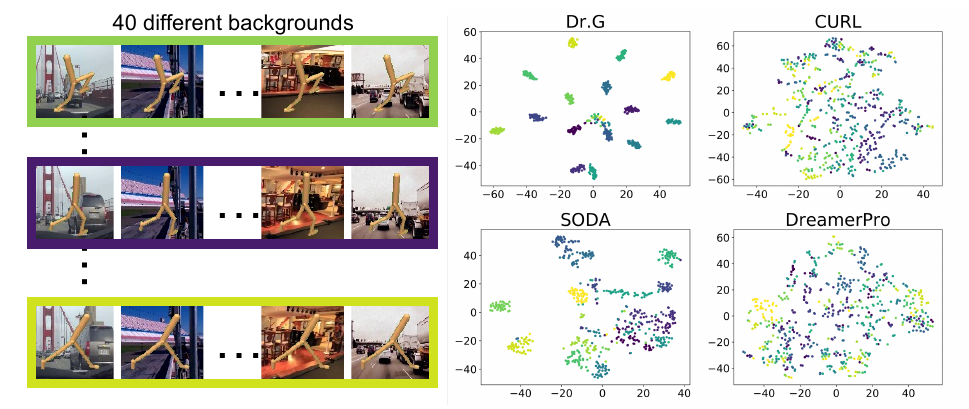}
\caption{Results of t-SNE of representations learned by \textbf{Dr. G}, CURL, SODA, and DreamerPro in the video hard setting. We randomly selected 40 backgrounds from the video hard and obtained t-SNE for about 15 motion situations. The color represents each motion situation, and each dot represents embedded latent for the same situation on a different background. Even when the background is dramatically different, \textbf{Dr. G} embeds behaviorally comparable data most closely.}
\label{fig:tsne}
\end{figure*}

\subsection{Baseline Methods}
We compare \textbf{Dr. G} with prior studies in both model-free and model-based algorithms. 
For model-free algorithms, we used the following as benchmarks: 
SAC~\cite{haarnoja2018soft}, which represents a straightforward soft actor-critic with no augmentation; 
CURL~\cite{laskin2020curl}, which involves applying a contrastive representation learning method; 
PAD~\cite{hansen2020self}, which represents SAC with inverse dynamics to fine-tune representation at test time; 
SODA~\cite{hansen2021generalization}, which involves learning representations by maximizing the mutual information between augmented and non-augmented data; 
and SECANT \footnote{The training code of SECANT is not open yet, and we brought some results from the SECANT paper.}~\cite{fan2021secant}, which is a self-expert cloning method that leverages image augmentation in two stages. 
We used the following as benchmarks for model-based algorithms: 
Dreamer~\cite{hafner2019dream}, which involves learning long-horizon behaviors by latent imagination with reconstruction;
and DreamerPro ~ \cite{deng2021dreamerpro}, which combines prototypical representation learning with temporal dynamics learning for a world model.


\subsection{DeepMind Control Suite}

The DeepMind Control suite is a vision-based simulator that provides a set of continuous control tasks. 
We experimented with six tasks; \textbf{ball in cup catch}, \textbf{cartpole swingup}, \textbf{cheetah run}, \textbf{hopper hop}, \textbf{walker run}, and \textbf{walker walk}. 
We used DMControl-GB~\cite{hansen2020self, hansen2021generalization} as a benchmark for vision-based reinforcement learning, which presents a challenging continuous control problem. 
All agents learned in the default environment (the background was fixed and the object to be controlled was placed on the skybox), as shown on the left of Figure~\ref{fig:train_test}. 
To evaluate the generalization performance to make an agent that can be applied to the real environment, we introduced two types of interference in the background. 
1) \textit{Video easy setting}: a relatively simple natural video (the dynamic of the background was small, as shown in the first row on the right of Figure~\ref{fig:train_test}~\cite{fan2021secant, hansen2021generalization, hansen2020self}). 
2) \textit{Video hard setting}: the distribution of disturbing factors changed dynamically and the skybox was removed~\cite{deng2021dreamerpro, hansen2021generalization, ma2020contrastive, nguyen2021temporal}, as shown in the second row on the right of Figure~\ref{fig:train_test}. 
As shown in Figure~\ref{fig:train_test}, we use Realestate10k and Kinetics400~\cite{kay2017kinetics} for testing on the disturbing background. Each RL method was trained for 500K environmental steps and was run with 3 seeds.

Table~\ref{result_table} shows that \textbf{Dr. G} achieved good generalization ability for the unseen observation changes on DeepMind Control, outperforming the baselines on 4 out of 6 tasks in the video easy setting and outperforming all baselines in the video hard setting. 
The first row of Table 1 shows the result of the evaluation in the video easy setting; \textbf{Dr. G} shows approximately 65\% improvement in generalization ability over the prior best-performing baseline. 
The second row of Table~\ref{result_table} is evaluated in the video hard setting, which includes large visual distribution shifts such as complex visual obstructions. 
The zero-shot generalization performance of \textbf{Dr. G} increased by 152\% over the state-of-the-art algorithms in all six environments. 
Except for SODA, all baseline methods show poor performance in the video hard setting. 
In Figure~\ref{fig:tsne}, we visualize the state embedding of the walker walk task using t-SNE~\cite{van2008visualizing}. 
A well-generalized agent should capture task-relevant (invariant) features when the image observations are behaviorally identical, even if the unseen backgrounds are significantly different. 
\textbf{Dr. G} can embed semantically similar observations most closely located in both the video easy and hard settings. 
This can lead to high zero-shot generalization performance, especially when the background changes include complex distractors unseen during training.
More experiment details are in the supplementary material.

\subsection{Robosuite}

\begin{figure}[t]
\centering
\includegraphics[width=0.42\textwidth]{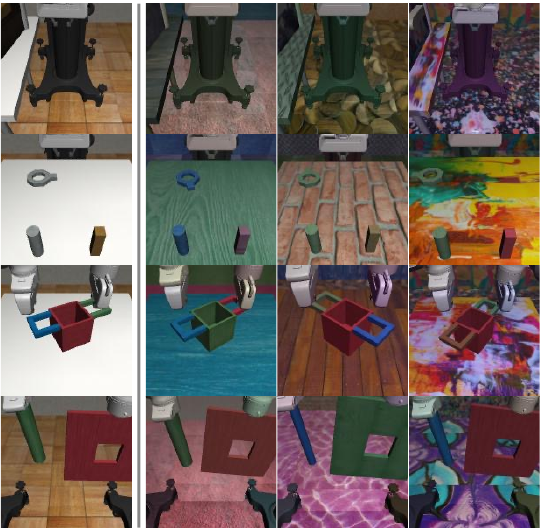}
\caption{Our agent is trained in a clean environment (first column from left) on Robosuite. We evaluate the ability to generalize in easy (second column), hard (third column), and extreme (fourth column) environments.}
\label{fig:robosuite}
\end{figure}

\begin{table}[t] 
\centering
\resizebox{0.49\textwidth}{!}{
\begin{tabular}{c|c|c|c|c}
    \toprule
    \textbf{Setting} & 
    \textbf{Task} & 
    \textbf{Dr. G} & 
    SODA & 
    DreamerPro \\
    \midrule
    \midrule
    \multirow{4}{*}{\textbf{\textit{Easy}}}
    & Door opening 
    & \textbf{465$\pm$26} 
    & 408$\pm$21 
    & 389$\pm$17 \\
    & Nut assembly  
    & 2.5$\pm$0.1 
    & \textbf{3.1$\pm$0.4}
    & 2.3$\pm$0.1 \\
    & Lifting
    & \textbf{432$\pm$23} 
    & 390$\pm$27
    & {335$\pm$16} \\
    & Peg-in-hole        
    & \textbf{320$\pm$28} 
    & 271$\pm$15 
    & 253$\pm$5 \\
    \midrule
    \multirow{4}{*}{\textbf{\textit{Hard}}}
    & Door opening 
    & \textbf{381$\pm$26} 
    & 368$\pm$19 
    & 341$\pm$25 \\
    & Nut assembly  
    & 1.8$\pm$0.5 
    & \textbf{2.8$\pm$0.6}
    & 1.7$\pm$0.7 \\
    & Lifting
    & \textbf{361$\pm$21} 
    & 323$\pm$27
    & {298$\pm$16} \\
    & Peg-in-hole        
    & \textbf{311$\pm$28} 
    & 245$\pm$15 
    & 211$\pm$5 \\
    \midrule
    
    \multirow{4}{*}{\textbf{\textit{Extreme}}}
    & Door opening 
    & \textbf{367$\pm$26} 
    & 331$\pm$19 
    & 307$\pm$25 \\
    & Nut assembly  
    & 1.9$\pm$0.4 
    & \textbf{3.3$\pm$0.6}
    & 2.9$\pm$0.2 \\
    & Lifting
    & \textbf{290$\pm$21} 
    & 266$\pm$17
    & {231$\pm$26} \\
    & Peg-in-hole        
    & \textbf{285$\pm$28} 
    & 238$\pm$15 
    & 203$\pm$5 \\
    \bottomrule
\end{tabular}
}
\caption{We trained models with default backgrounds on four tasks in Robosuite, and evaluated them on different background settings; Easy, Hard, and Extreme. Each task was run with 3 seeds.}
\label{result_table2}
\end{table}

Robosuite~\cite{zhu2020robosuite} is a modular simulator for robotic research. 
We benchmarked \textbf{Dr. G} and other methods on two single-arm manipulation tasks and two challenging two-arm manipulation tasks.
We used the Panda robot model with operational space control and trained with task-specific dense rewards. 
All agents received image observations from an agent-view camera as input. 
\textbf{Door opening}: a robot arm must turn the handle and open the door in front of it. 
\textbf{Nut Assembly}: a robot must fit the square nut onto the square peg and the round nut onto the round peg.
\textbf{Two Arm Lifting}: two robot arms must grab a handle and lift a pot together, above a certain height while keeping the pot level.
\textbf{Two Arm Peg-In-Hole}: two robot arms are placed next to each other. One robot arm holds a board with a square hole in the center, and the other robot arm holds a long peg. The two robot arms must coordinate to insert the peg into the hole. 
All agents were trained with clean backgrounds and objects like the first column in Figure~\ref{fig:robosuite}. 
We evaluated generalization performance in three unseen visual distractors environment.
The three environments in Figure~\ref{fig:robosuite} are easy (second column), hard (third column), and extreme (fourth column).
Each RL method was trained for 500K environment steps and was run with 3 seeds.

Table~\ref{result_table2} lists the results of the evaluation. 
We compared it to other four algorithms; SODA~\cite{hansen2021generalization}, DreamerPro~\cite{deng2021dreamerpro}, CURL~\cite{laskin2020curl},  and Dreamer~\cite{hafner2019dream} as baselines. Due to space limitations, only SODA and DreamerPro, which perform better, are presented.
\textbf{Dr. G} achieves better generalization performance than other models in all environments with unseen visual distractions, except the Nut-assembly environment, which implies that \textbf{Dr. G} is suitable for real-world deployment for robotic manipulation.

\subsection{Ablation Studies}
We performed four ablation studies to evaluate the importance and synergy of each component of \textbf{Dr. G}. 
We evaluated the absence of DCL and RSID, the combination of soft-hard augmentation, the type of hard augmentation, and the relationship between baseline and hard augmentation.
Here, an ablation study is introduced to show the difference in performance according to each module. For the other three ablation studies, refer to the supplementary material.

\begin{figure}[t]
\centering
\includegraphics[width=0.49\textwidth]
{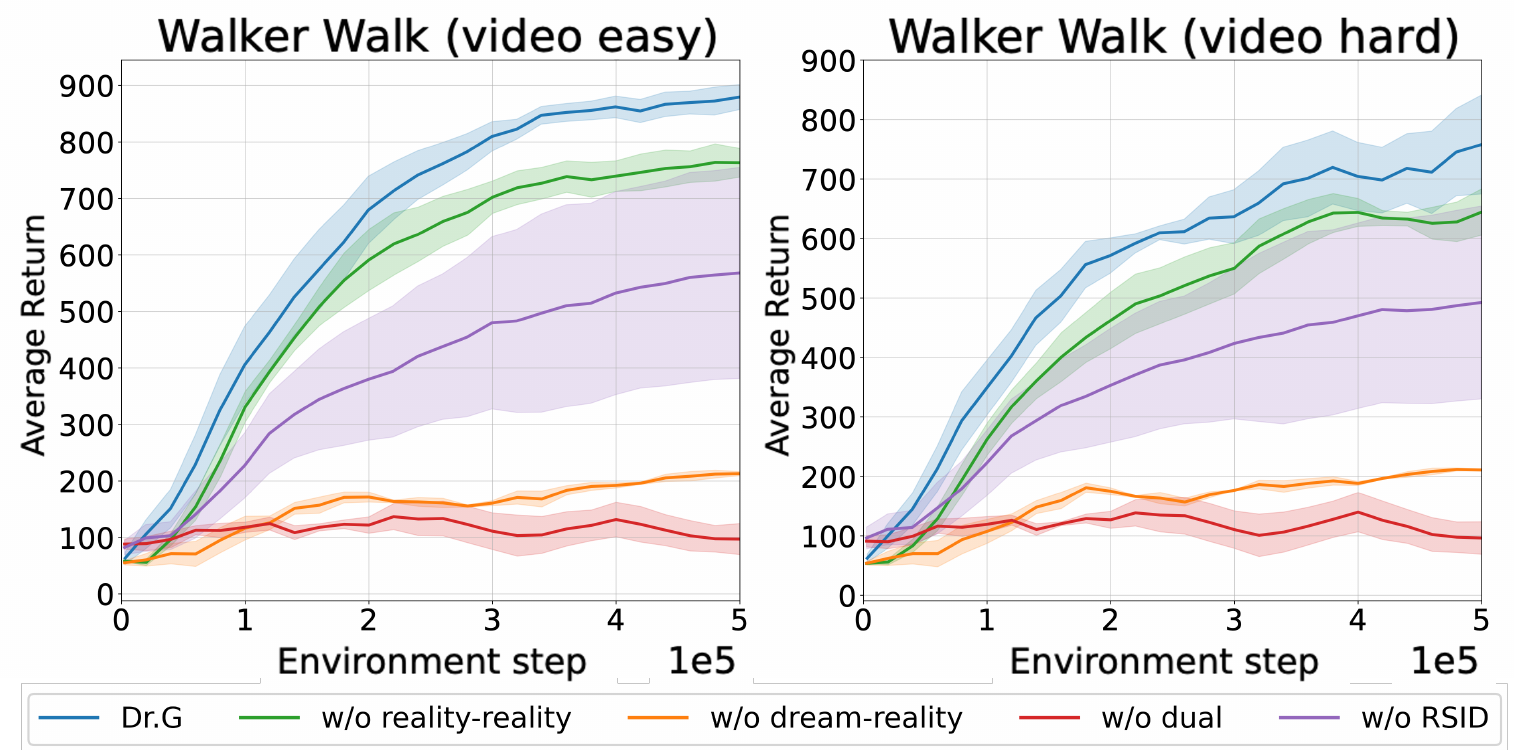}
\caption{Ablation study on effects of each module in \textbf{Dr. G}. Each task was run with 3 seeds.}
\label{fig:ablation2}
\end{figure}

\subsubsection{Effects of Each Module}
We analyzed the individual effects of each module: \textit{dual contrastive} (\textit{dream-reality} and \textit{reality-reality}), and \textit{RSID}; results are shown in Figure \ref{fig:ablation2}. 
We removed one of the modules for \textit{w/o dream-reality}, \textit{w/o reality-reality}, and \textit{w/o RSID}.
Specifically, we removed all dual contrast objectives for \textit{w/o dual}.
In \textit{w/o dual} and \textit{w/o dream-reality}, their performance was very poor. 
Because \textbf{Dr. G} eliminates reconstruction loss, dream-reality contrastive learning is essential for training the world model. 
In the case of \textit{w/o RSID}, it shows that the zero-shot performance degrades considerably, as shown in Figure~\ref{fig:ablation2}.

\section{Conclusion} 
\label{sec:conclusion}
In this study, we proposed \textbf{Dr. G}, a novel self-supervised learning method for zero-shot MBRL in visual control environments. 
The proposed encoder and world model are trained by a combination of DCL and RSID over two-way data augmentation. 
We demonstrated the generalization performance of \textbf{Dr. G} in the DeepMind control suit. After training with standard (simple and clean) backgrounds, we test \textbf{Dr. G} with unseen visual distractions. We also showed the visual randomizing tests in a realistic robot manipulation simulator, Robosuite. Through the extensive simulation results, \textbf{Dr. G} demonstrates the best zero-shot generalization performance compared to existing model-based and model-free RL methods.

\section*{Acknowledgements}
This work was supported partly by the Institute of Information and Communications Technology Planning and Evaluation (IITP) grant funded by the Korea Government (MSIT) (No. 2022-0-01045, Self-directed Multi-Modal Intelligence for solving unknown, open domain problems), (No. 2022-0-00688, AI Platform to Fully Adapt and Reflect Privacy-Policy Changes), and (No. 2019-0-00421, Artificial Intelligence Graduate School Program(Sungkyunkwan University)).

\bibliography{aaai23.bib}

\end{document}